# Guidelines for Artificial Intelligence Containment

**James Babcock**
Cornell University
jab299@cornell.edu

**Janos Kramar**
University of Montreal
jkramar@gmail.com

**Roman V. Yampolskiy**
University of Louisville
roman.yampolskiy@louisville.edu

**Abstract**
With almost daily improvements in capabilities of artificial intelligence it is more important than ever to develop safety software for use by the AI research community. Building on our previous work on AI Containment Problem we propose a number of guidelines which should help AI safety researchers to develop reliable sandboxing software for intelligent programs of all levels. Such safety container software will make it possible to study and analyze intelligent artificial agent while maintaining certain level of safety against information leakage, social engineering attacks and cyberattacks from within the container.

## 1. Introduction

The past few years have seen a remarkable amount of attention on the long-term future of AI. Icons of science and technology such as Stephen Hawking[1], Elon Musk[2], and Bill Gates[3] have expressed concern that superintelligent AI may wipe out humanity in the long run. Stuart Russell, coauthor of the most-cited textbook of AI [35], recently began prolifically advocating[4] for the field of AI to take this seriously. AI conferences now frequently have panels and workshops on the topic[5,6]. There has been an outpouring of support[7] from many leading AI researchers for an open letter calling for greatly increased research dedicated to ensuring that increasingly capable AI remains "robust and beneficial", and gradually a field of "AI safety" is coming into being [64, 33, 55]. Why all this attention?

Since the dawn of modern computing, the possibility of artificial intelligence has prompted leading thinkers in the field to speculate [15, 43, 46] about whether AI would end up overtaking and replacing humanity. However, for decades, while computing quickly found important applications, artificial intelligence remained a niche field, with modest successes, making such speculation seem irrelevant. But fast-forwarding to the present, machine learning has seen grand successes and very substantial R&D investments, and it is rapidly improving in major domains, such as natural language processing and image recognition, largely via advances in deep learning [18]. Artificial *General* Intelligence (AGI), with the ability to perform at a human-comparable level at most cognitive tasks, is likely to be created in the coming century; most predictions, by both experts and non-experts, range from 15-25 years [7].

As the state of research in AI capabilities has steadily advanced, theories about the behavior of superintelligent AGIs have remained largely stagnant, though nearby scientific fields examining optimal agents (game theory and decision theory), idealized reasoning (Bayesian statistics and the formal theory of causality), and human cognition (cognitive neuroscience) have come into being and given us some clues. A prominent theory is that an advanced AI with almost any goals will generically develop certain subgoals called "basic AI drives" [31], such as self-preservation, self-improvement, and resource acquisition. Pursuit of these goals could motivate the AI to, for example, make copies of itself on internet-connected computers, build new hardware or software for itself, and evade the attention of human observers until it is confident that it's beyond their control. An influential book [10] thoroughly reviewing and building on existing work on superintelligent AI found no compelling counterarguments or easy workarounds; to the best of our knowledge, safe AGI will require significant theoretical advances in AI safety, and very careful implementation. This implies a risk that the first human-level AGI will be unsafe, at least until research and testing are done on it.

---

[i] AI Boxing, AI Containment, and AI Confinement are assumed to mean the same thing.
[1] http://www.bbc.com/news/technology-30290540
[2] http://video.cnbc.com/gallery/?video=3000286438
[3] https://www.reddit.com/r/IAmA/comments/2tzjp7/hi_reddit_im_bill_gates_and_im_back_for_my_third/
[4] http://www.cse.unsw.edu.au/~tw/aiethics/AI_Ethics/Introduction.html
[5] https://people.eecs.berkeley.edu/~russell/research/future/aamas14-workshop.html
[6] http://ijcai13.org/program/special_events_the_impact_of_ai_on_society
[7] http://futureoflife.org/ai-open-letter

## 1.1 Containment and the AI Safety Problem

Due to the "basic AI drives" mentioned above, an unsafe AGI will likely be motivated to falsify tests or monitoring mechanisms to manipulate the researchers into thinking it's safe, to gain access to more resources, to embed dormant copies of itself in device firmwares, and to hack computers on the internet. In order to reliably test and safely interact with an AGI with these motivations and capabilities, there must be barriers preventing it from performing these actions. These barriers are what we refer to as containment.

Some have argued that controlling AGI - especially if superintelligent - is impossible or infeasible. For example, Ray Kurzweil writes that "intelligence is inherently impossible to control" [17]. Eliezer Yudkowsky's AI box experiment[8] found that human factors make containing an AI difficult. Vernor Vinge argued that "confinement is intrinsically impractical" in the long run [45].

We agree that containment is not a long-term solution for AI safety; rather it's a tool to enable testing and development of AGIs with other, more robust safety properties such as value learning [36, 67] and corrigibility [37]. Value learning is the strategy of programming an AGI to learn what humans value, and further those values. If this is done correctly, such an AGI could be very good for humanity, helping us to flourish. Corrigibility is the strategy of programming an AGI to help (or at least, to not resist) its creators in finding and fixing its own bugs. An AGI which had both of these properties would not need to be contained, but experience with software suggests that developers are very unlikely to get it right on the first try.

Other safety strategies that have been proposed depend on containment more directly. For example, in his book Superintelligence [10], Nick Bostrom suggests using tripwires to monitor the AGI and shut it down if it appears to be behaving dangerously. However, the AI drives thesis [31] suggests that an AGI might try to bypass tripwires or remove them from itself, which would render them ineffective in an AGI that had full control over its hardware. On the other hand, an AI containment system with internal security boundaries could both keep an AI from disabling its tripwires, and keep it from learning the details of what tripwires there were.

Regarding the tractability of containment, encouraging progress has been made on the human factors front; Yampolskiy has proposed ways of limiting an AGI's communication channels so that even a superintelligent AI could not trick an operator into doing something dangerous [54, 6]. As for preventing the AGI from tampering with data and hacking its way to the internet, essentially no work has been done on this problem, but we have reason to think that bringing the tools of cybersecurity to bear will yield results that will substantially mitigate the risk of escapes.

## 2. Overview of the Proposed Guidelines

The proposed guidelines are based on our analysis of AI containment, incorporating all aspects of the problem, a concrete technical perspective and attention to specific technologies that can be used[*]. Background research for developing this framework turned up surprising interactions between seemingly disparate aspects of the problem. For example, existing research into the human factors of AI containment suggests that tripwire/intrusion detection systems

---

might need to censor details of what happened. Understanding the threat model requires incorporating widely varied predictions, and these predictions lead to very different ways of approaching the security/usability tradeoff [4].

We have identified seven major subproblems to the AI containment. Rather than focusing our attention narrowly on one subproblem, we have made it our priority not to have any blind spots. These subproblems are threat modeling, navigating the security/usability tradeoff, concrete mechanisms, sensitive information management, human factors, tripwires and graceful degradation. This last issue - the need for graceful degradation - has led us to envision containment systems with multiple dissimilar containment mechanisms combined with intrusion detection, providing defense in depth and resilience in a system built from components which may individually contain bugs.

**A prototype AI container** A prototype AI container cab be built by combining multiple pre-existing containment tools, in such a way that a security vulnerability in any one of them would not lead to a complete compromise of the system. Developers should provide source code, scripts, and documentation sufficient for others to easily use it. Additionally, it is important to test how the container interacts with the most common machine learning tools, including, in particular, those that involve distributed computation and GPGPU computing, and should have working support for at least some GPGPU and distributed computing mechanisms.

**Analysis of the AI containment threat model** A preliminary threat model is included in this paper. To produce a more comprehensive threat model, would require a literature review of AI safety and forecasting results that may affect the threat model. It is necessary to attempt to determine upper and lower bounds on an AGI's ability to find specific types of security vulnerabilities at different levels of intelligence and information constraint.

**Analysis of the security/usability tradeoff** Our guidelines define what constitutes light, medium, and heavy containment. Future efforts will survey available options within each category, and analyze risk at each level of containment in terms of the threat model to produce necessary guidance for AGI developers' risk analysis, indicating which level of containment is likely to be necessary. The details of how options are classified will depend mainly on how the options interact with the types of computations and development tools used, for which present-day machine learning practices can be used as a proxy for the sorts of tools likely to be used by AGI developers in the future.

**Limits of provably-secure communication** Proposed guidelines propose provably-secure communication protocols, suitable for use in a "heavy containment" context where security means preventing an AGI from communicating messages complex enough to be used for social engineering attacks. It will be necessary to explore the information theoretic limits of how much information can be communicated while maintaining maximum security.

**New Categories of Sensitive Information Created by AGI Development** AGI development creates new categories of sensitive information. Some information is sensitive in the sense that it should be kept away from the AGI; some information is sensitive in that an AGI could use it as a tool for social engineering attacks. Our guidelines suggest identifying in greater detail what types of information are sensitive, and provide guidance on how to manage it.

**Analysis of Human Factors** It is necessary to analyze the channels in which an AGI might be able to place messages for untrained personnel, and identify actions which create unintuitive risks that people near an AGI should be warned about. AI safety researchers need to explore methods of preventing disallowed AGI communication with untrained personnel, such as understanding all possible side channels, by adding noise to remove steganographically hidden messages from intended outputs, and by controlling the ways output is viewed.

**Analysis of AI Container Tripwires** AI safety researchers will need to explore the space of feasible tripwires, creating a guide for implementers describing what tripwires would be desirable and how they would be implemented.

## 3. Significance of AI Containment

This research sits at the intersection between cybersecurity, machine learning, and the emerging field of AI safety. AI forecasting determines our threat model; we need predictions about how intelligent future AIs are likely to be and the software, hardware, and organizational contexts in which it might be created. These are not straightforward questions which can be simply looked up; predictions in the literature feature a wide variety of possible AI scenarios, motivations, and types, leading to an equal variety of overlapping threat models. Practices within the field of machine learning determine the shape of the usability-security tradeoff. Existing security tools and techniques can help, but only with reanalysis in light of this new threat model.

Our guidelines bring AI containment from the abstract theoretical domain, where it previously resided, into the cybersecurity world with a threat model, software and procedures that are concrete enough to check for vulnerabilities.

### 3.1 Why Present Techniques Aren't Adequate

Futurists have speculated on such extreme strategies as choosing computations which make processors trigger unknown corner cases in physics, placing messages in log files which hack the brains of humans who read them, or reprogramming non-radio hardware components to act as radios by creating and detecting RF interference [16]. These attacks may seem unlikely, but a more pressing concern comes from simply searching the Common Vulnerabilities and Exposures (CVE) database for entries related to common virtualization and sandboxing technologies. We found that every major virtualization and sandboxing solution we looked at had critical vulnerabilities discovered within the past two years, with vulnerabilities continuing to be found at a rate which suggests that there are many more vulnerabilities left to be discovered. While it is probably feasible to find and fix all of the vulnerabilities that are easy enough for a human to find, there is no reason to think that any of these solutions could stand up to a smarter-than-human intelligence nor that they will ever be strong enough to stand up to a smarter-than-human intelligence.

One might ask: why not just use airgapping? We are mainly concerned with scenarios involving AGIs being tested under laboratory conditions, where physical disconnection from the internet is possible. Unfortunately, this does nothing to protect the integrity of tests being performed, to protect monitoring systems, or to protect the hardware being used from modification and persistence, which are all very necessary.

### 3.2 Why Now

Given that AGI has not been created yet, one might ask: why now? There are several compelling reasons. Human-level AGI may be invented sooner than expected. Technology forecasting is notoriously difficult, and has tripped up serious scientists time and again; for example the nuclear physicist Ernest Rutherford said in 1933 that nuclear energy was "moonshine", and another accomplished physicist Lord Kelvin in 1896 said "I have not the smallest molecule of faith in aerial navigation other than ballooning".

Another reason to do this work now is that developing secure systems requires time, careful review by many independent experts, and iteration. Unfortunately, many of the predicted scenarios do not allow second chances; if an AI that's unfriendly to human values or that's dangerously buggy escapes into the world, it could do immense damage. If we wait until someone has created a human-level AGI, or the creation of human-level AGI appears imminent, it will be too late. Some of this research has an even longer lead time; if our research finds security improvements that depend on changing software in ways more complicated than mere bugfixes, then finding out now will allow time to influence the next generation of software.

A working AI container will improve the security of the internet and the world in any future scenarios where artificial general intelligence is developed, enabling prospective AGI designs which might not share human values to be safely studied and assessed. It will also bridge the cybersecurity, AI safety and machine learning communities. The direct application of our research is very important, but more speculatively, approaching security from this unconventional angle may yield insights into old problems, particularly those related to dealing with risk from newly discovered exploits and highly reliable systems.

## 4. Previous Work

To succeed in constructing an AI container it is important to merge techniques from a number of relevant fields such as: computer security [52], behavioral biometrics [61, 51], cryptography [22, 63], steganography [1], stylometry [3, 2], computer forensics [21, 9], utilization of artificial intelligence in security [30, 29], AI Safety [23, 28], including security work with current AIs [48] and theoretical work on Artificial General Intelligence (AGI) safety and security [49].

Other relevant work is in the field of Artimetrics, which deals with ways to identify, classify, and authenticate robots, intelligent software, and virtual reality agents for security purposes [60, 47]. Other tools may include linguistic profiling of Chatbots [3, 2], the visual identification of virtual reality avatars [27, 62, 24], and the behavior-based authentication of bots [62]. More recent areas of importance are AI Safety Engineering [50], AGI Safety [41], and utility function security [58], AI-Completeness [56], the singularity paradox [59], the space of minds [57], recursive self-improvement [53], and the Artificial Intelligence Containment Problem (AICP) [54]. In particular, Yampolskiy's seminal work on the AICP not only defines the problem, from the computer science point of view, but also proposes a number of possible solutions, including a steganographically-safe communication protocol, a classification of levels of communication with respect to the amount of shared information, and the concept of safe questions for interacting with AGIs. In the same work Yampolskiy also analyzes a number of potential attack vectors against the confinement cyberinfrastructure, including social engineering attacks.

## 5. Evaluation and Success Criteria

**Soundness** Schneier's Law says that "anyone can invent a security system that he himself cannot break." For a containment system that's meant to stand up to smarter-than-human adversaries, that's not good enough. In the long run, good security can only be guaranteed by time and careful review by many independent experts [34]; allowing time for review and iteration is the main reason to do this research now, when AI may still be decades away, rather than waiting until it's closer. It is commonly said that perfect security is impossible, but when the opponent might wield superhuman abilities, it's important to get very close.

It is important to design a containment system that's sound in the sense that we'll be confident in its security *eventually*, after sufficient independent review. Needing a few simple bugfixes is okay; presenting something that won't ever be secure even after said bugfixes is not.

**Breadth** There is a lot of uncertainty about how AGI may first arise; eg whether it will be developed inside a company or in academia, what size the team will be, what sorts of algorithms and technologies it will use. As much as possible, it is important to cover all reasonable scenarios, and find all the likely attack surfaces and mitigations.

**Relevance and Usability** The next major risk is that these guidelines will produce something that's secure but unusable in practice, or worse, something which is only usable up to a point, after which an AGI researcher would break something security-critical to work around a limitation. There are two main ways this could happen in practice. The first is incompatibility with the tools and libraries that AI researchers are using. The second is overly obnoxious configuration.

In order to learn and pragmatically respond to the priorities and practical constraints of cutting-edge machine learning researchers, AI container developers need to establish collaborations with AI researchers. Such collaborations will be needed to ensure that the containment measures do not unnecessarily interfere with existing tools and research practices, by finding out what they're using, testing it for compatibility, and distributing container prototypes for external evaluation. When soliciting feedback, it is important to engage with all the major subcommunities within AI, such as the "Good Old Fashioned AI" community and the machine learning community.

## 6. Threat Model

### 6.1 AI Motivation

The "AI drives" thesis gives us a starting point for reasoning about what an AGI might want; according to the thesis, while researchers may give the AGI any goal explicitly, it will likely develop subgoals like self-defense, self-improvement, self-replication, and resource acquisition. This allows us to recognize what AGI behaviors are likely: those that further these subgoals. Details of how this scenario would play out is explored in greater depth in Nick Bostrom's book *Superintelligence* [10]. On the other hand, if the thesis is false, it may still be helpful as a sort of upper bound on the difficulty of the problem; after all, what could be more dangerous than a superintelligence bent on world domination?

**6.2 AI Capabilities**

Predictions about AGI development trajectories tend to fall into two main categories, known as *hard takeoff* and *soft takeoff* [10]. In a *soft takeoff*, the capabilities of AGI increase gradually, with much time and effort; and it will be feasible to learn about its capabilities and behaviors through experimentation, though as Yudkowsky [66] points out, the change in capability as the AGI surpasses us may enable new plans and behaviors that are undesirable.

In a so-called *hard takeoff*, after some critical point is reached (which may be preceded by steady improvement), AGI capability increases very rapidly. One proposed mechanism for this is that AGIs that are as good at AGI research as their researchers can undergo a process of *recursive self-improvement* [15, 39, 20, 32], where each improvement in an AGI's capability enables it to find more improvements to make. Whether a hard or soft takeoff will happen is a subject of considerable disagreement among experts[9] [68], but the possibility of recursive self-improvement makes an AGI's ability to self-modify particularly important to monitor and control.

As far as the ceiling of AI capability is concerned, it's likely to be significantly above human level in general [19, 40], though it's not clear what hardware is required for this. Of course, this raises the question: what do we mean by a smart AGI, anyway? As commentators often point out, a superhuman chess player like Deep Blue is no match for us at any other tasks, even simple ones like tic-tac-toe. For present purposes, we will answer this question using a list of "cognitive superpowers" humans have, proposed by Bostrom [10]:

- "**Intelligence amplification** (AI programming, cognitive enhancement research, social epistemology development, etc.)
- **Strategizing** (Strategic planning, forecasting, prioritizing, and analysis for optimizing chances of achieving distant goals)
- **Social manipulation** (Social and psychological modeling, manipulation, rhetoric persuasion)
- **Hacking** (Finding and exploiting security flaws in computer systems)
- **Technology research** (Design and modeling of advanced technologies, such as biotechnology or nanotechnology)
- **Economic productivity** (Various skills enabling economically productive intellectual work)"

Using this framework we can say that:

- An AGI with strong strategizing and hacking ability might plan an escape using security vulnerabilities;
- An AGI with strong technology ability might additionally be able to find surprising side-channel attacks
- An AGI with strong social manipulation ability could use social engineering techniques to enable an escape;
- An AGI with other superpowers may additionally be able to make compelling offers or threats, which could assist with the manipulation.

Countermeasures for these attack vectors are discussed in the remaining sections.

## 7. The Security-Usability Tradeoff

Existing work on the AGI containment problem [54] has generally assumed that the AGI in question is superintelligent, and therefore warrants a great deal of concern: any exploit that seems theoretically possible will be within the AGI's reach, and therefore the only acceptable containment system is one that we have strong theoretical reason to believe is immune to circumvention. This is a high bar for a containment system to clear; we'll call containment that we can be confident meets this standard **heavy containment**. Some examples of containment mechanisms that belong in this category:

- Limiting output to a small number of yes/no questions whose answers are securely shuffled before being output [54], to avoid any chance of social engineering attacks

---

[9] http://reducing-suffering.org/predictions-agi-takeoff-speed-vs-years-worked-commercial-software

- Running the AGI as a homomorphically encrypted computation[10]
- Putting the AGI's datacenter in a very inaccessible place

But as we noted in the Threat Model section, researchers developing AGI are unlikely to be willing to conduct their research with such substantial handicaps. Researchers who believe their project is aimed at a breakthrough short of human-comparable AGI are particularly likely to balk at costs during use and during setup, and will only be willing to deploy containment mechanisms that are made simple and convenient. We'll call these **light containment**. It's unlikely that light containment measures will suffice for a superintelligent AGI, but there may also be containment escape risks from less powerful AGIs, and effective light containment would help mitigate those risks. Light containment is particularly relevant to groups that don't yet have AGIs at all, and only need to ensure that their AGI doesn't escape before they notice they've made a breakthrough. Technologies like hardware virtualization, network firewalls, and software sandboxing belong in this category.

Researchers who understand that they're dealing with potentially human-level AGI will probably be able to deploy containment measures that add modest costs and overheads to their development process, but in a competitive scenario as in [5] heavy containment measures will likely be out of bounds; they would cause anyone applying them to fall behind. Measures that keep costs within the same order of magnitude, however, will have a good chance of being taken by responsible researchers. We'll call measures in this category **medium containment**. Possible measures in this category include:

- Preventing the AI from accumulating information about its researchers between runs
- Adding real-time external redundant supervision, which is tested using appropriately varied drills
- Instruction-level virtualization
- Specialized hardware

If reliable containment mechanisms are developed that don't impose significant computational overhead but are incompatible with the development environments that the researchers would otherwise find most productive, then those will be best thought of as medium containment, until the compatibility issues are addressed. This underscores the importance of investigating the requirements and techniques in advance, to forestall this sort of situation.

Existing AI safety research has focused primarily on measures suitable for heavy containment - that is, measures with high reliability but with a very high cost to computational performance and AGI functionality. This is sensible; containing a superintelligence is very hard, and people's risk and difficulty estimates are easily miscalibrated, and it is more important to have containment than it is to have it run fast and support the best development tools. Still, this should not be the end of the story. These guidelines are the first to meaningfully address light containment and medium containment. It aims to strike an appropriate balance between caution and practicality.

## 8. Sensitive Information Management

In most applications of cybersecurity, there isn't very much uncertainty about which kinds of information are sensitive and need to be restricted; but when an AGI enters the picture, this becomes much less clear, because an AGI that's trying to evade containment may be helped by information about what stands in its way. Detailed information about this might comprise personal information about the researchers, books on psychology, negotiation or salesmanship, or books on cybersecurity and manuals to the software and hardware the AGI is running on. Any of these would make it easier for the AGI to invent attacks. A slightly more sophisticated AGI would be able to make use of more indirect information, such as works of literature to learn about psychology or content about computing, engineering, or physics to learn about its environment. However, there's no fundamental reason an AGI would be limited to learning from such explicit information; an AGI that can perform experiments may be able to figure out its environment the hard way, as humanity has.

Limiting access to detailed information about the AGI's containment is superficially similar to what is referred to pejoratively as "security through obscurity", but there is an important difference between the usual cybersecurity context and the AGI containment context: researchers have control over all of the information that the contained AGI learns from. For this reason, hiding information from the AGI is more feasible than maintaining secrets in a deployed system.

---

[10] http://lesswrong.com/lw/3cz/cryptographic_boxes_for_unfriendly_ai

This kind of information restriction is palatable for many applications of AGI. However, it poses a challenging tradeoff with the requirements of value learning, which is another key project for AGI safety; value learning requires information about our preferences (stated or enacted, abstract or concrete, etc.), and this information would be relevant to manipulating people. This tradeoff will require more research attention; there may be adequate subsets of preference information for developing value learning techniques that are nonetheless inadequate for social engineering attacks. From our current vantage point it seems that the requirements of value learning will diminish the effectiveness of information-hiding measures targeted at obstructing social engineering, so more intensive non-information-hiding-based measures will be needed to manage the resulting risk.

## 8.1 Strategies

The light containment approach is to avoid giving access to sensitive information of the kinds mentioned. Because this is a broad class of data, there would also be a risk of some researcher accidentally providing sensitive data that's embedded in some larger dataset; a possible light containment tool would be something akin to a content filter, though ultimately discretion about what data is shared would necessarily fall to the researcher.

A medium containment approach would be to prevent the AGI from deducing that it's running as a particular piece of software in the world by letting it interact only with a virtual environment, or some computationally well-defined domain, with as few embedded clues as possible about the outside world. (Arguably projects like the DeepMind's Atari-game-playing AI [25] are already doing this.)

If it was possible to embed all of the AGI's computations (not merely its action, reward, and sense streams) inside a virtual world and get theoretical guarantees that a full description of the virtual world doesn't contain enough information to deduce anything significant about the world outside, then this would be a way of providing heavy containment. Getting these theoretical guarantees may not be tractable, however.

The most pressing aspects of the AGI Containment problem are to open all avenues of investigation, and to conduct research on subtopics that require a lot of lead time before their results can be realistically put into practice, usually due to technical complexity. The sensitive information management problem does not appear to have very much technical depth, unless we consider the heavy containment problem, which seems mathematically challenging. However, AI safety researchers will need to bring clarity to the question of what constitutes sensitive information.

This topic has been discussed in an AI safety context, though not, to our knowledge, investigated deeply. There is substantial uncertainty about whether our intuitions about what can be learned from a corpus are an acceptable guide; for example, Yudkowsky[11] has speculated that a very small amount of data is enough to reveal much of how the world works to a superintelligent mind, while Dileep George, cofounder of the AI startup Vicarious, has argued that a great deal of gradual observation is needed for knowledge acquisition. Some theoretical work has been done on the topic, but it's not sufficient to answer practical questions [38].

## 9. Concrete Mechanisms

There are many off-the-shelf tools that implement sandboxing. However, none of them were developed with the AI containment threat model in mind and many of them are difficult to reconcile with the requirements posed by AI software development. We've grouped these into categories; sandboxing techniques within a category tend to have similar strengths and limitations, and it is generally more useful to combine techniques from different categories than from the same category. A prototypical AI container will attempt to combine techniques from as many distinct categories as is feasible, in a modular fashion. The categories we have identified are:

- **Physical isolation**, aka airgapping.

---

[11] http://lesswrong.com/lw/qk/that_alien_message

- **Language-semantics based methods**, such as the Java sandbox - i.e., the use of a sound type-system with memory safety to prevent the introduction of new code. Of particular interest are Lua (with the Torch7 library) and Python (with the Theano library), which might be able to implement this kind of sandbox but which incorporate components in C that might make them insecure. These are currently popular software packages for implementing machine learning algorithms [11].
- **Operating-system-based process sandboxing methods**, such as Linux's seccomp mode, FreeBSD's Capsicum, and the Chrome sandbox. These work by starting a process and then signaling to the operating system that most APIs will not be used and should be disabled, before the execution of any risky code.
- **Virtual machines**, such as VirtualBox and QEMU. These use CPU hardware features to create multiple access levels within a computer, each with its own operating system. These might be improved by new security features in future CPU architectures, such as the Mill[12], which has features to defend against some classes of exploits.
- **Instruction-level emulation**. This technique is used heavily in malware analysis tools such as V2E [65], but imposes heavy computational overhead.

Since many software sandboxing tools already exist, we think creating another one from scratch is likely to be premature. Instead, AI safety researchers should concentrate on finding existing tools, attempting to adapt them to the AI containment threat model and analyzing them in that context.

The main difficulty will be adapting these sandboxing methods to a multi-node cluster context; AGIs are expected to depend heavily on parallel computation, which means they need to run on multiple hosts connected to each other. This will most likely mean a cluster of compute nodes within a single data center, connected to each other but not to the internet, where there is both a high-privilege enforcer and a low-trust AGI process on each node. Since none of these sandboxing mechanisms were designed with this requirement in mind, we expect most of these mechanisms to be either unsuitable, or to require considerable adaptation and configuration, performed with great care.

The other major consideration will be the need for GPGPUs computing. Because GPGPU involves using two heterogeneous compute engines and the interaction between them, it tends to require significantly greater privileges than CPU computation. In particular, GPU drivers combine a very complicated attack surface with an unavoidable need for low-level hardware access, posing considerable risk. We are not yet sure how best to address this risk, but it might be possible to shrink this attack surface by validating and limiting GPU API calls or modifying open source graphics drivers.

## 10. Graceful Degradation

### 10.1 Defense in Depth

Because an AGI could potentially discover novel security vulnerabilities, it's important for an AGI containment system to remain secure even if there are vulnerabilities in its components. Therefore, we envision a system featuring defense in depth, with multiple techniques providing overlapping guarantees of integrity. Conceptually, these overlapping techniques act as layers, in the sense that the attack surfaces of an outer layer are only accessible to probing and attack if the integrity guarantees of all inner layers have been broken. This should be combined with intrusion detection systems, so that any attempt by an AI to break the integrity constraints imposed on it would be noticed, allowing the developers to shut it down. Ideally the traps and detection systems would be implemented in higher layers, making them more difficult to detect and circumvent. When deciding which interfaces to expose and which interfaces to architect around protecting, researchers should look at design attributes like complexity and check exploit databases such as NVD[13], OSVDB[14], and EDB[15], using projects' security history as a warning if they're likely to have a bad security future.

---

[12] http://millcomputing.com/topic/security/
[13] https://nvd.nist.gov
[14] http://osvdb.org
[15] http://exploit-db.com

This part of our research applies to all levels of containment (light, medium, and heavy); a medium containment system would use a subset of the layers used by a heavy containment system, while a light containment system would use a subset of the layers used by a medium containment system. These guidelines will enable developers to set up effective containment systems, assess their systems' security, and assess the consequences of a breach or partial breach. Our initial work in this area has indicated which features of AGI containment systems are most in need of work, and we expect continued work in this area to expose more necessary lines of inquiry.

## 11. Tripwires

Tripwires are systems that monitor the operation of a running AGI, and shut it down if they detect an anomaly that suggests the AGI might be malfunctioning or unsafe. For example, one might install a tripwire which monitors the AGI's thoughts for signs that it was planning to deceive its developers [10], or a tripwire which monitors the AGI's execution for signs that it had self-modified or self-improved. Similarly, intrusion detection systems are tripwires which detect if code is executing somewhere it shouldn't; they increase the value of multi-layer architectures like the one described in the previous section, by adding a cost to unsuccessful attempts to break the system's integrity and preventing an AI from keeping its progress in breaking out over time.

We classify tripwires into two general categories. *Semantic* tripwires refer to the contents of an AGI's thoughts or plans; their design requires detailed knowledge of the algorithms and data structures it will be using. *Technical* tripwires do not refer to the AGI's thoughts or plans, but only to algorithm-independent aspects of its execution. In this classification, detecting that the AGI was formulating a plan to deceive its developers would be a semantic tripwire, while detecting that the AGI had self-modified would be a technical tripwire. Intrusion detection systems are another kind of tripwire.

Semantic tripwires present highly unusual requirements. Conceptually, a semantic tripwire is similar to the *assert* statement used in many programming languages. Unlike a simple *assert* statement, however, it's desirable to keep the implementation details of a semantic tripwire inaccessible to the program it's located in; it would improve security if the tripwires could be somehow excluded from the AI's readable address space. Because the algorithms that will be used by real AGIs have mostly not been developed yet, designing the specific contents of semantic tripwires does not yet seem feasible; AI safety researchers should instead focus on designing technical tripwires, and on algorithm-independent aspects of tripwire placement, such as how to add tripwires in such a way that an AGI with unrestricted read access to its own memory couldn't study them to find workarounds. AI safety researchers should also investigate the human factors involved in tripwires, which could be ignored or disabled by AGI developers if they seem to be having false positives, to disastrous effect.

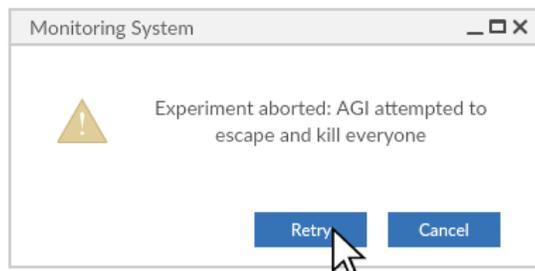

**Figure 3.** User error can negate most defenses

There are many intrusion detection systems [26, 13] which might be adapted to AI containment. However, they assume a significantly different threat model and context which may limit their applicability and which requires great attention to detail in order to operate. Existing research into honeypots [42] will also be helpful for designing tripwired/fake escape paths.

## 12. Human Factors

AI containment presents a mix of both conventional human-factor problems, and some very novel and unusual ones. The most conventional problem is that for anyone involved in a scenario involving AGI, there are very high stakes, which can lead to high stress and heightened risk of mistakes. There are several problems that are specific to the circumstances, but reminiscent of similar problems seen elsewhere. There is a risk that an employee could release the AGI deliberately, for ideological reasons [12, 14]. There is also a risk that external parties could undermine the security of the containment system: for example, an AGI development lab with a working prototype is likely to attract espionage.

The more unusual risk comes from the AGI itself. An AI with human-level social manipulation powers could have many ways to use social engineering to escape, and AI with other cognitive powers (or superhuman social manipulation powers) could muster significant leverage to negotiate with. An AGI which had superhuman intelligence would be quite persuasive indeed.

What's more, just as an AGI might use side-channels or advanced persistent threats to break its technical containment infrastructure, it may also find similarly difficult-to-control ways to convey messages to the researchers or to distant third parties, eg by adjusting its computation to shape its power-usage graphs or embedding steganographically hidden messages in images it produces - presumably other modes also exist that we're not yet aware of. The actions that release the AI can look harmless or unrelated, especially if messages reach non-experts who don't understand the significance of technical actions like plugging things in.

Most existing security human-factors research involves preventing problematic actions by laypeople or by people who may be spread out through a large company. In the case of AI containment, only very a small number of people need to interact directly with an AGI, and so they can be screened and trained carefully, making the problem easier. On the other hand, those few carefully selected personnel could be exposed to extremely clever trickery.

Informal studies of the human factors of AI containment have provided reason to think it's difficult. The AI box experiment [69] and replication [44] provide data about the sorts of strategies an AGI would use to convince humans to let it out of a container deliberately, but countermeasures have not yet been explored, nor have the strategies one might use to make humans open a container accidentally.

## 13. Conclusions

In the face of the challenges of developing and controlling increasingly sophisticated AIs, it is crucial to anticipate and restrict the ways an AI could perform unwanted actions. There has been little work on establishing practical ways for AI research and development to enforce these restrictions; our paper provides guidelines so that the technologies and information required will be ready to be used by AI projects as they advance toward human-comparable intelligence and beyond.

### Acknowledgements

Authors are grateful to Jaan Tallinn and Effective Altruism Ventures for providing funding towards this project, and to Victoria Krakovna and Evan Hefner for their feedback. Roman Yampolskiy is also grateful to Elon Musk and the Future of Life Institute for partially funding his work.

### References

[1] G. Abboud, J. Marean and R. V. Yampolskiy, *Steganography and Visual Cryptography in Computer Forensics*, *Systematic Approaches to Digital Forensic Engineering (SADFE), 2010 Fifth IEEE International Workshop on*, IEEE, 2010, pp. 25-32.

[2] N. Ali, M. Hindi and R. V. Yampolskiy, *Evaluation of Authorship Attribution Software on a Chat Bot Corpus*, *23rd International Symposium on Information, Communication and Automation Technologies (ICAT2011)*, Sarajevo, Bosnia and Herzegovina, October 27-29, 2011, pp. 1-6.

[3] N. Ali, D. Schaeffer and R. V. Yampolskiy, *Linguistic Profiling and Behavioral Drift in Chat Bots*, Midwest Artificial Intelligence and Cognitive Science Conference (2012), pp. 27.

[4] D. Amodei, C. Olah, J. Steinhardt, P. Christiano, J. Schulman and D. Mané, *Concrete problems in AI safety*, arXiv preprint arXiv:1606.06565 (2016).

[5] S. Armstrong, N. Bostrom and C. Shulman, *Racing to the precipice: a model of artificial intelligence development*, AI & society, 31 (2016), pp. 201-206.

[6]    S. Armstrong, A. Sandberg and N. Bostrom, *Thinking inside the box: Controlling and using an oracle ai*, Minds and Machines, 22 (2012), pp. 299-324.

[7]    S. Armstrong and K. Sotala, *How we're predicting AI–or failing to*, *Beyond Artificial Intelligence*, Springer, 2015, pp. 11-29.

[8]    J. Babcock, J. Kramar and R. Yampolskiy, *The AGI Containment Problem*, *The Ninth Conference on Artificial General Intelligence (AGI2015)*, NYC, USA, July 16-19, 2016.

[9]    M. B. Beck, E. C. Rouchka and R. V. Yampolskiy, *Finding Data in DNA: Computer Forensic Investigations of Living Organisms*, *Digital Forensics and Cyber Crime*, Springer Berlin Heidelberg, 2013, pp. 204-219.

[10]   N. Bostrom, *Superintelligence: Paths, dangers, strategies*, Oxford University Press, 2014.

[11]   R. Collobert, K. Kavukcuoglu and C. Farabet, *Implementing neural networks efficiently*, *Neural Networks: Tricks of the Trade*, Springer, 2012, pp. 537-557.

[12]   E. Dietrich, *After the Humans are Gone*, Journal of Experimental & Theoretical Artificial Intelligence, 19(1) (2007), pp. 55-67.

[13]   T. Garfinkel and M. Rosenblum, *A Virtual Machine Introspection Based Architecture for Intrusion Detection*, *Ndss*, 2003, pp. 191-206.

[14]   H. D. Garis, *The Artilect War*, ETC publications, 2005.

[15]   I. J. Good, *Speculations Concerning the First Ultraintelligent Machine*, Advances in Computers, 6 (1966), pp. 31-88.

[16]   M. Guri, A. Kachlon, O. Hasson, G. Kedma, Y. Mirsky and Y. Elovici, *Gsmem: Data exfiltration from air-gapped computers over gsm frequencies*, *24th USENIX Security Symposium (USENIX Security 15)*, 2015, pp. 849-864.

[17]   R. Kurzweil, *The Singularity is Near: When Humans Transcend Biology*, Viking Press, 2005.

[18]   Y. Lecun, Y. Bengio and G. Hinton, *Deep learning*, Nature, 521 (2015), pp. 436-444.

[19]   S. Legg, *Machine Super Intelligence*, *PhD Thesis*, *University of Lugano*, Available at: http://www.vetta.org/documents/Machine_Super_Intelligence.pdf, June 2008.

[20]   R. Loosemore and B. Goertzel, *Why an intelligence explosion is probable*, *Singularity Hypotheses*, Springer, 2012, pp. 83-98.

[21]   M. Losavio, O. Nasraoui, V. Thacker, J. Marean, N. Miles, R. Yampolskiy and I. Imam, *Assessing the Legal Risks in Network Forensic Probing*, Advances in Digital Forensics V (2009), pp. 255-266.

[22]   A. Majot and R. Yampolskiy, *Global catastrophic risk and security implications of quantum computers*, Futures, 72 (2015), pp. 17-26.

[23]   A. M. Majot and R. V. Yampolskiy, *AI safety engineering through introduction of self-reference into felicific calculus via artificial pain and pleasure*, *IEEE International Symposium on Ethics in Science, Technology and Engineering*, IEEE, Chicago, IL, May 23-24, 2014, pp. 1-6.

[24]   R. Mcdaniel and R. V. Yampolskiy, *Embedded non-interactive CAPTCHA for Fischer Random Chess*, *16th International Conference on Computer Games (CGAMES)*, IEEE, Louisville, KY, 2011, pp. 284-287.

[25]   V. Mnih, K. Kavukcuoglu, D. Silver, A. A. Rusu, J. Veness, M. G. Bellemare, A. Graves, M. Riedmiller, A. K. Fidjeland and G. Ostrovski, *Human-level control through deep reinforcement learning*, Nature, 518 (2015), pp. 529-533.

[26]   C. Modi, D. Patel, B. Borisaniya, H. Patel, A. Patel and M. Rajarajan, *A survey of intrusion detection techniques in cloud*, Journal of Network and Computer Applications, 36 (2013), pp. 42-57.

[27]   A. Mohamed, N. Baili, D. D'souza and R. V. Yampolskiy, *Avatar Face Recognition Using Wavelet Transform and Hierarchical Multi-scale LBP*, *Tenth International Conference on Machine Learning and Applications (ICMLA'11)*, Honolulu, USA, December 18-21, 2011.

[28]   L. Muehlhauser and R. Yampolskiy, *Roman Yampolskiy on AI Safety Engineering*, *Machine Intelligence Research Institute*, Available at: http://intelligence.org/2013/07/15/roman-interview/ July 15, 2013.

[29]   D. Novikov, R. V. Yampolskiy and L. Reznik, *Anomaly Detection Based Intrusion Detection*, *Third International Conference on Information Technology: New Generations (ITNG 2006)*, Las Vegas, Nevada, USA, April 10-12, 2006., pp. 420-425.

[30]   D. Novikov, R. V. Yampolskiy and L. Reznik, *Artificial Intelligence Approaches for Intrusion Detection*, *Long Island Systems Applications and Technology Conference (LISAT2006).* , Long Island, New York., May 5, 2006., pp. 1-8.

[31]   S. M. Omohundro, *The Basic AI Drives*, *Proceedings of the First AGI Conference, Volume 171, Frontiers in Artificial Intelligence and Applications, P. Wang, B. Goertzel, and S. Franklin (eds.)*, IOS Press, February 2008.

[32]   S. M. Omohundro, *The Nature of Self-Improving Artificial Intelligence*, *Singularity Summit*, San Francisco, CA, 2007.

[33]   F. Pistono and R. V. Yampolskiy, *Unethical Research: How to Create a Malevolent Artificial Intelligence*, *25th International Joint Conference on Artificial Intelligence (IJCAI-16). Ethics for Artificial Intelligence Workshop (AI-Ethics-2016)*, 2016.

[34]   B. Potter and G. Mcgraw, *Software security testing*, IEEE Security & Privacy, 2 (2004), pp. 81-85.

[35]   S. Russell and P. Norvig, *Artificial Intelligence: A Modern Approach*, Prentice Hall, Upper Saddle River, NJ, 2003.

[36]   N. Soares, *The value learning problem*, Machine Intelligence Research Institute, Berkley, CA, USA (2015).

[37]   N. Soares, B. Fallenstein, S. Armstrong and E. Yudkowsky, *Corrigibility*, *Workshops at the Twenty-Ninth AAAI Conference on Artificial Intelligence*, Austin, Texas, USA, January 25-30, 2015.

[38]   R. J. Solomonoff, *A formal theory of inductive inference. Part I*, Information and control, 7 (1964), pp. 1-22.

[39] R. J. Solomonoff, *The Time Scale of Artificial Intelligence: Reflections on Social Effects*, North-Holland Human Systems Management, 5 (1985), pp. 149-153.

[40] K. Sotala, *Advantages of artificial intelligences, uploads, and digital minds*, International Journal of Machine Consciousness, 4 (2012), pp. 275-291.

[41] K. Sotala and R. V. Yampolskiy, *Responses to catastrophic AGI risk: a survey*, Physica Scripta, 90 (2015), pp. 018001.

[42] L. Spitzner, *Know your enemy: Honeynets*, Honeynet Project (2005).

[43] A. M. Turing, *Intelligent Machinery, A Heretical Theory*, Philosophia Mathematica, 4(3) (1996), pp. 256-260.

[44] Tuxedage, *I attempted the AI Box Experiment again! (And won - Twice!)*, Available at: http://lesswrong.com/lw/ij4/i_attempted_the_ai_box_experiment_again_and_won/, September 5, 2013.

[45] V. Vinge, *The Coming Technological Singularity: How to Survive in the Post-human Era*, *Vision 21: Interdisciplinary Science and Engineering in the Era of Cyberspace*, Cleveland, OH, March 30-31, 1993, pp. 11-22.

[46] N. Wiener, *Cybernetics or Control and Communication in the Animal and the Machine*, MIT press, 1961.

[47] R. Yampolskiy, G. Cho, R. Rosenthal and M. Gavrilova, *Experiments in Artimetrics: Avatar Face Recognition*, Transactions on Computational Science XVI (2012), pp. 77-94.

[48] R. Yampolskiy and J. Fox, *Safety engineering for artificial general intelligence*, Topoi, 32 (2013), pp. 217-226.

[49] R. V. Yampolskiy, *Artificial intelligence safety engineering: Why machine ethics is a wrong approach*, *Philosophy and Theory of Artificial Intelligence*, Springer, 2013, pp. 389-396.

[50] R. V. Yampolskiy, *Artificial Superintelligence: a Futuristic Approach*, Chapman and Hall/CRC, 2015.

[51] R. V. Yampolskiy, *Behavioral Modeling: an Overview*, American Journal of Applied Sciences, 5 (2008), pp. 496-503.

[52] R. V. Yampolskiy, *Computer Security: from Passwords to Behavioral Biometrics*, New Academic Publishing, 2008.

[53] R. V. Yampolskiy, *From Seed AI to Technological Singularity via Recursively Self-Improving Software*, arXiv preprint arXiv:1502.06512 (2015).

[54] R. V. Yampolskiy, *Leakproofing Singularity - Artificial Intelligence Confinement Problem*, Journal of Consciousness Studies (JCS), 19(1-2) (2012), pp. 194–214.

[55] R. V. Yampolskiy, *Taxonomy of Pathways to Dangerous Artificial Intelligence*, *Workshops at the Thirtieth AAAI Conference on Artificial Intelligence*, 2016.

[56] R. V. Yampolskiy, *Turing Test as a Defining Feature of AI-Completeness*, *Artificial Intelligence, Evolutionary Computation and Metaheuristics - In the footsteps of Alan Turing. Xin-She Yang (Ed.)*, Springer, 2013, pp. 3-17.

[57] R. V. Yampolskiy, *The Universe of Minds*, arXiv preprint arXiv:1410.0369 (2014).

[58] R. V. Yampolskiy, *Utility function security in artificially intelligent agents*, Journal of Experimental & Theoretical Artificial Intelligence, 26 (2014), pp. 373-389.

[59] R. V. Yampolskiy, *What to Do with the Singularity Paradox?*, *Philosophy and Theory of Artificial Intelligence*, Springer, 2013, pp. 397-413.

[60] R. V. Yampolskiy and M. L. Gavrilova, *Artimetrics: Biometrics for Artificial Entities*, Robotics & Automation Magazine, IEEE, 19 (2012), pp. 48-58.

[61] R. V. Yampolskiy and V. Govindaraju, *Behavioural biometrics: a survey and classification*, International Journal of Biometrics, 1 (2008), pp. 81-113.

[62] R. V. Yampolskiy and V. Govindaraju, *Strategy-based behavioural biometrics: a novel approach to automated identification*, International Journal of Computer Applications in Technology, 35 (2009), pp. 29-41.

[63] R. V. Yampolskiy, J. D. Rebolledo-Mendez and M. M. Hindi, *Password Protected Visual Cryptography via Cellular Automaton Rule 30*, *Transactions on Data Hiding and Multimedia Security IX*, Springer Berlin Heidelberg, 2014, pp. 57-67.

[64] R. V. Yampolskiy and M. Spellchecker, *Artificial Intelligence Safety and Cybersecurity: a Timeline of AI Failures*, arXiv preprint arXiv:1610.07997 (2016).

[65] L.-K. Yan, M. Jayachandra, M. Zhang and H. Yin, *V2E: combining hardware virtualization and softwareemulation for transparent and extensible malware analysis*, ACM Sigplan Notices, 47 (2012), pp. 227-238.

[66] E. Yudkowsky, *Artificial Intelligence as a Positive and Negative Factor in Global Risk*, in N. Bostrom and M. M. Cirkovic, eds., *Global Catastrophic Risks*, Oxford University Press, Oxford, UK, 2008, pp. 308-345.

[67] E. Yudkowsky, *Complex value systems in friendly AI*, Artificial general intelligence, Springer, 2011, pp. 388-393.

[68] E. Yudkowsky and R. Hanson, *The Hanson-Yudkowsky AI-foom debate*, *MIRI Technical Report*, Available at: http://intelligence.org/files/AIFoomDebate.pdf, 2008.

[69] E. S. Yudkowsky, *The AI-Box Experiment*, Available at: http://yudkowsky.net/singularity/aibox, 2002.